\documentclass[oneside,psamsfonts,reqno,a4paper,final,english]{article}
\usepackage[includehead, includefoot, centering]{geometry}

\usepackage[utf8]{inputenc} 
\usepackage[T1]{fontenc}    
\usepackage{hyperref}       
\usepackage{url}            
\usepackage{booktabs}       
\usepackage{amsfonts,amsmath,amssymb}       
\usepackage{nicefrac}       
\usepackage{microtype}      

\usepackage{algorithm}
\usepackage[noend]{algpseudocode}
\usepackage{stmaryrd}
\usepackage{graphicx}
\usepackage{epsfig}
\usepackage{color}

\usepackage[toc,page]{appendix}

\title{A Connection between Feed-Forward Neural Networks and Probabilistic Graphical Models}
\date{}

\author{
  Dmitrij Schlesinger\\
  Dresden University of Technology
  \thanks{This work was supported by German Federal Ministry of Education and Research (BMBF, 01IS14014A-D). The computations were performed on an HPC Cluster at the Center for Information Services and High Performance Computing (ZIH) at TU Dresden.}
}

\begin{document}

\maketitle

\begin{abstract}
Two of the most popular modelling paradigms in computer vision are feed-forward neural networks (FFNs) and probabilistic graphical models (GMs). Various connections between the two have been studied in recent works, such as e.g.~expressing mean-field based inference in a GM as an FFN. This paper establishes a new connection between FFNs and GMs. Our key observation is that any FFN implements a certain approximation of a corresponding Bayesian network (BN). We characterize various benefits of having this connection. In particular, it results in a new learning algorithm for BNs. We validate the proposed methods for a classification problem on CIFAR-10 dataset and for binary image segmentation on Weizmann Horse dataset. We show that statistically learned BNs improve performance, having at the same time essentially better generalization capability, than their FFN counterparts.
\end{abstract}

\section{Introduction}
In this work we study connections between feed-forward neural networks (FFNs) and graphical models (GMs). The latter can be divided in two classes -- directed and undirected ones (see e.\,g.~\cite{Koller:2009}). The corresponding statistical models are known as Bayesian networks (BNs) and Markov random fields (MRFs), respectively. On the neural network side, perhaps the two most popular types are restricted Boltzmann machines (RBMs, \cite{ackley1985learning}) and feed-forward networks. RBMs are statistical models, i.\,e.~they implement Boltzmann distributions, which are nothing but MRFs. Hence, there is a one-to-one correspondence between RBMs and undirected GMs. The situation is quite different for FFNs. Although the architectures of FFNs and BNs are basically the same, FFNs do not represent probability distributions at all. They are considered just as deterministic mappings, transforming observed values into output values. The main aim of this work is to bridge this gap between FFNs and BNs by showing that each FFN implements an approximation of a corresponding BN.

Undirected GMs have been used successfully for decades, especially in computer vision. Directed GMs appear to be less popular. At the same time, deep learning methods have found tremendous success in recent years, especially convolutional neural networks (CNNs, \cite{lecun1998gradient}). The most successful techniques, however, have no statistical background as of yet. Hence from a practical viewpoint, the goal of this work is to establish a connection that makes a wide range of elaborate statistical inference and learning techniques available to modern deep architectures.

\paragraph{Related works.} There are many works showing how certain algorithms for graphical models can be represented as applications of specially designed networks. One popular approach is e.\,g.~to express mean-field based inference techniques as neural networks, which allows for very fast implementations \cite{Kr:NIPS11}. In \cite{zheng2015conditional} it is shown how to express mean-field based inference as a differentiable layer in a CNN. In \cite{lin2015deeply} it is proposed to directly learn messages for energy minimization, again expressed in terms of neural computations. In these works, however, only specific aspects are considered. The full power of MRFs, including uncertainty, statistical learning, decisions other than maximum a-posteriori etc.~is not exploited. More importantly, the connections established in these works go \emph{from GM to FFN}, i.\,e.~for a particular GM algorithm, the corresponding FFN architecture is designed. Our aim, in contrast, is to address the opposite direction, i.\,e.~to be able to transform a given FFN to the corresponding GM, thereby taking into account that the latter implements a complex probability distribution, rather than an optimization problem.

The observation that incorporating stochastics into the learning process can improve performance is not new. The most famous example may be the ``drop-out'' method \cite{srivastava2014dropout}. Injecting noise (see e.\,g.~\cite{BengioLC13} and references therein) is also a popular approach. Note how the use of randomness improves the generalization capabilities of the learned models, as it is usual for statistical models in contrast to discriminative ones. However, these approaches use stochastics in a rather ad-hoc manner, i.\,e.~the proposed procedures are not derived from a statistically sound probability model for all constituents.

Probably the closest to our work are \cite{Saul:1996} and \cite{TangS13}, where algorithms for learning Bayesian networks are presented. The approach in \cite{Saul:1996} employs techniques commonly used for undirected graphical models. In particular, the Bayesian network is transformed into the corresponding undirected graphical model of higher order, and an adapted version of mean-field approximation is applied to it. Finally, the algorithm uses a complex message passing scheme, which, while very interesting from a theoretical point of view, seems to be not efficient enough for very deep models and large data. The algorithm in \cite{TangS13} in contrast does not rely on the mean-field approximation but aims to maximize the original likelihood using importance sampling. As acknowledged by the authors, importance sampling has certain requirements in order to work in practice, namely that the proposal distribution should be ``not small''. Unfortunately, for very deep architectures, it seems to be just the opposite case. Note that in both of these works no connection between FFN and BN is explicitly given. In other words, there are no statements on how to interpret {\em deterministic} computations performed by an FFN in terms of a corresponding {\em statistical} BN. In contrast, our work provides this interpretation, making it possible to directly compare FFNs to their statistical counterparts.

From the viewpoint of modeling, our work is motivated by the following observation. One of the recent trends in computer vision is to combine FFNs and conditional random fields (CRFs, \cite{lafferty2001conditional}) in a unified framework, and to learn them jointly (see e.g.~\cite{arnabconditional}). There are different ways to achieve this. For instance, a CRF can be seen as a differentiable layer in an FFN as in \cite{zheng2015conditional}. Another option is to use a FFN for computing the potentials of a CRF sitting on top of it. As an example, in \cite{ChenSYU15} a maximum likelihood based learning scheme is presented, where the gradient of the likelihood is approximately computed using belief propagation. In \cite{accv2016} an efficient sampling-based approach is presented. However, in both \cite{ChenSYU15} and \cite{accv2016} the FFN is again only considered as a mapping, producing CRF potentials from the input in a deterministic manner. There is no statistical model or interpretation of the FFN as such. Obviously, for such FFN-CRF combinations it would be profitable to be able to represent both FFN and CRF in a unified manner, because then they could be combined in a transparent way and the same learning principles applied for both parts.

\paragraph{Contributions.} First, we show that for each FFN, i.e.~a {\em deterministic mapping}, a corresponding probabilistic directed graphical model, or Bayesian network, can be defined that implements a {\em probability distribution} such that the former is an approximation of the latter. We discuss possible benefits from having such a correspondence with respect to modeling, learning and inference. 

Second, based on this correspondence, we derive a learning algorithm for BNs, which has the following properties. On the one side, it implements a statistical learning. As a consequence, learned probabilistic models are less vulnerable to over-fitting as compared to the corresponding discriminative/deterministic approaches. At the same time, unlike many statistical learning algorithms, our method enjoys perfect sampling, that makes it quite efficient in practice. On the other side, the proposed method is very similar to the commonly used error back-propagation (EBP) algorithm for FFNs. It potentially allows the proposed approach to use numerous findings made for FFNs in the past years, like e.g.~elaborated general solvers. Nonetheless, due to its similarity to the EBP, the proposed method can be seamlessly incorporated into existing deep learning frameworks, like e.g.~Caffe \cite{jia2014caffe}.

In the experimental section, we evaluate the proposed approach with respect to the obtained performance and generalization capability for a classification task on CIFAR-10 database and for binary segmentation on Weizmann Horse dataset. We show improved performance and great improvement of the generalization capability of the proposed method as compared to the corresponding deterministic models.


\section{Statistical interpretation of FFNs}
\paragraph{A unified representation of different types of neurons.} We start from considering a simple sigmoid neuron: $y = \text{sigm}(\langle x, w\rangle)$, where $\text{sigm}(a) = \exp\bigl(a\bigr) / \bigl(1+\exp(a)\bigr)$, $x$ is the input vector, and $w$ are the weights. Its statistical interpretation is well known to be a logistic regression. Let us introduce a random variable $y \in \{0,1\}$ and its conditional probability distribution $p(y|x;w) \propto \exp(y \cdot \langle x,w\rangle)$ (log-linear model). Then $\text{sigm}(\langle x,w\rangle)=p(y{=}1|x;w)$. Hence, the outputs of sigmoid neurons in an FFN can be interpreted as probability values. 

The problems that prevent us from following this specific interpretation of FFN neurons are the following. For one, the idea can not be applied to many other types of neurons, such as $\tanh$ or RELU, simply because these neurons compute values that may be negative or unbounded. But we deem the next problem to be even more crucial. Assume that the output of a sigmoid neuron serves as input for another one. Then this output value is multiplied by a weight, which can be arbitrary. This operation, however, has no statistical interpretation, since the obtained value can have arbitrary range depending on the weight. The values $x$ for a neuron are usually understood as features. But it is hardly possible to reinterpret probability values as features, as it is difficult to pose a well-founded statistical model with clearly defined sample space, random variables, probability distribution etc.~in this case.

The key idea is to interpret the output of a sigmoid neuron in an FFN not as a probability value, but as an expectation\footnote{For now, we assume that the values of $y$ are elements of a vector space, i.e.~addition and multiplication with scalar are well-defined. Further explanation can be found in appendix \ref{appendix_a}.}
\begin{equation}\label{eq:sigmexp}
	\text{sigm}(\langle x,w\rangle) = \mathbb E_{p(y|x;w)}[y] ,
\end{equation}
with $y$ and $p(y|x;w)$ as described above. Now it is easy to see how this interpretation can be applied for other neuron types. Consider e.\,g.~a $\tanh$ neuron with the transfer function
\begin{equation}\label{eq:tanh}
	\tanh(a) = \frac{\exp(a)-\exp(-a)}{\exp(a)+\exp(-a)} .
\end{equation}
Let us introduce the random variable $y \in \{-1,+1\}$ and the conditional probability distribution $p(y|x;w) \propto \exp(y \cdot \langle x,w\rangle)$. Then $\tanh(\langle x,w\rangle) = \mathbb E_{p(y|x;w)}[y]$, similar to \eqref{eq:sigmexp}.

Concerning the RELU neuron, we borrow the idea of \cite{nair2010rectified}, where it is shown that RELUs can be represented as a sum of sigmoid neurons with appropriately defined weights. Since the sum of expectations is the expectation of a sum, we can assert the applicability of \eqref{eq:sigmexp} for RELUs as well.

In fact, it is possible to represent any deterministic mapping $y = f(x)$ as an expectation. The simplest variant is
\begin{equation}\label{eq:delta}
	f(x) = \mathbb E_{p(y|f(x))}[y] \ \ \text{with}\ \ p(y|f(x)) = \llbracket y = f(x)\rrbracket ,
\end{equation}
where $\llbracket\cdot\rrbracket$ returns $1$ if its argument is true. Of course, this interpretation is not very useful, since the function remains in fact deterministic -- there is only one possible value of $y$ for each value of $f(x)$. It asserts, however, the possibility to represent {\em all thinkable} neuron types (e.\,g.~pooling, normalization etc.) in a unified manner.

To summarize, neurons can be represented in a unified manner by means of the following constituents: (i) two random variables $x$ and $y$, and (ii) a conditional probability distribution $p(y|x)$. According to this representation, the application of a neuron consists in the computation of the expectation of $y$,
\begin{equation}\label{eq:nappr}
	\bar y(x) = \mathbb E_{p(y|x)}[y] .
\end{equation}


\paragraph{FFN and the corresponding BN.} Consider an FFN that consists of neurons $V = \{v_1\ldots v_n\}$. Each neuron has its inputs $I(v) \subset V$. The simplest way to represent an FFN is as a directed graph $G = (V,E)$, where nodes correspond to neurons and directed edges are defined as $E = \{(u,v)|u\in I(v)\}$. Each neuron $v$ is associated with its output value $z_v$ and is characterized by a function $z_v = f_v(z_{I(v)})$, where $z_A$ denotes the vector of values for all neurons $v\in A\subset V$. Some neurons $v\in X \subset V$ are input ones, namely those for which $I(v) = \emptyset$. The values of these neurons are not computed by the network, but are just given as observations. There are also output neurons $v\in Y \subset V$, i.\,e.~those which do not serve as input for some other neurons in the network. The set of all other neurons is denoted by $Z = V \setminus (X \cup Y)$. The feed-forward architecture means that the graph $G = (V,E)$ has no directed cycles. Hence, it is possible to introduce a partial ordering $\succeq$ on the set of neurons, such that each neuron has higher rank than its inputs, i.\,e.~$u \in I(v) \Rightarrow v \succ u$. The application of the network consists of computing the output values $z_Y$ given the input values $z_X$. Since the neurons are ordered, the computation of $z_Y$ can be performed sequentially according to the introduced partial ordering $\succeq$.

The corresponding probabilistic directed graphical model, or Bayesian network, is built as follows\footnote{Here, we use similar notation with slightly different meaning. For example, $z_v$ in an FFN denotes the value computed by the neuron $v$, whereas in a BN, $z_v$ denotes a random variable or its value, depending on context.}. First, we introduce random variables for each neuron. Let us denote by $x_v, v \in X$ variables that correspond to the input neurons, $y_v, v \in Y$ are output variables, and $z_v, v \in Z$ are all intermediate variables. Each variable takes values that correspond to the particular type of neuron. For example, $z_v \in \{0,1\}$ if $v$ is a sigmoid neuron, $z_v \in \{-1,+1\}$ if $v$ is a $\tanh$ neuron, $x_v \in \mathbb R$ for input neurons etc. In order to unify and simplify notation, let us also introduce ``unified input vectors'' for all neurons (analogous to the previously introduced $z_{I(v)}$): Let $I_v$ be a vector of input values for a neuron $v$. These values can be $x$ or $z$ for both $v \in Z$ and $v \in Y$. For each non-input neuron $v \in (Y \cup Z)$, we define the conditional probability distribution $p(z_v|I_v)$ or $p(y_v|I_v)$, which also corresponds to a particular type of neuron (e.\,g.~$p(z_v|I_v;w) \propto \exp(z_v \cdot \langle I_v,w\rangle)$ for both sigmoid neuron and $\tanh$, where $w$ are the weights). An elementary event in our statistical model is a vector $(\mathbf x,\mathbf y, \mathbf z)$ of values for all variables, where $\mathbf x$ denotes the vector of all input values; $\mathbf y$ and $\mathbf z$ defined analogously. The model is the conditional probability distribution
\begin{equation}\label{eq:prob}
p(\mathbf y, \mathbf z|\mathbf x) = p(\mathbf z|\mathbf x)\cdot p(\mathbf y|\mathbf z,\mathbf x)
= \prod_{v\in Z} p(z_v|I_v) \cdot \prod_{v \in Y} p(y_v|I_v) .
\end{equation}
In order to be inline with graphical model representation, let us define clique functions as $f_v(z_v,I_v) = -\ln p(z_v|I_v)$ for all $v \in Z$ and $f_v(y_v,I_v) = -\ln p(y_v|I_v)$ for all $v \in Y$. Then the probability distribution \eqref{eq:prob} can be written as $p(\mathbf y, \mathbf z|\mathbf x) = \exp(-E(\mathbf x, \mathbf y, \mathbf z))$ with the energy
\begin{equation}\label{eq:gm}
	E(\mathbf x, \mathbf y, \mathbf z) = \sum_{v\in Z} f_v(z_v,I_v) + \sum_{v\in Y} f_v(y_v,I_v) .
\end{equation}
Note that unlike in undirected GMs, there is no normalization constant $Z$, since we use logarithms of real conditional probabilities as clique functions.


\paragraph{FFNs approximate the corresponding BNs.} For simplicity, let us consider first a BN as described above with only one hidden layer, i.e.~$I_v$-s for all $v\in Z$ contain only input variables $x_v$. Let also the output $\mathbf y$ depends only on $\mathbf z$, i.e.~$p(\mathbf y|\mathbf z,\mathbf x)=p(\mathbf y|\mathbf z)$ for clarity. Since $z$-s are auxiliary variables, at the end we are interested in the conditional probability distribution of the output variables given the input, i.e.~$p(\mathbf y|\mathbf x)$, which is obtained by marginalization over auxiliary variables $\mathbf z$ by $p(\mathbf y|\mathbf x)=\sum_{\mathbf z} p(\mathbf y|\mathbf z)\cdot p(\mathbf z|\mathbf x)$. This probability distribution can be seen as a convex combination of probability distributions $p(\mathbf y|\mathbf z)$ with different values of the argument $\mathbf z$ and coefficients $p(\mathbf z|\mathbf x)$. Let us approximate the convex combination of function values by the value of the function, obtained at convex combination of its argument, i.e. 
\begin{equation}\label{eq:meanfield}
p(\mathbf y|\mathbf x)=\sum_{\mathbf z} p(\mathbf y|\mathbf z)\cdot p(\mathbf z|\mathbf x)\approx p(\mathbf y|\bar{\mathbf z}(\mathbf x)) \text{\ , where\ \ \ } \bar{\mathbf z}(\mathbf x)=\sum_{\mathbf z} \mathbf z\cdot p(\mathbf z|\mathbf x) .
\end{equation}
Since $p(\mathbf z| \mathbf x)$ is conditionally independent (see \eqref{eq:prob}), the components of $\bar{\mathbf z}(\mathbf x)$ are expectations
\begin{equation}
\bar{z}_v(\mathbf x)=\sum_{z_v} z_v\cdot p(z_v|\mathbf x)=\mathbb E_{p(z_v|\mathbf x)}[z_v] .
\end{equation}
This is however exactly what neurons compute (see \eqref{eq:nappr}). To conclude, FFN with one hidden layer approximates the true probability distribution $p(\mathbf y|\mathbf x)$ of the corresponding BN using \eqref{eq:meanfield}.

It is very easy to see how to extend the above statement to the models with more than one hidden layer. Let us consider the layered structure as follows. We partition the set of intermediate variables $v\in Z$ into $L$ layers according to the previously introduced order $\succeq$ of neurons (i.e.~inputs $I_v$ for neurons in the $l$-th layer belong to previous layers $l'<l$). Let us denote by $\mathbf z_l$ the vector of all variable values in the $l$-th layer. The probability distribution \eqref{eq:prob} can be rewritten as\footnote{Here we also assume that variables in the $l$-th layer depend only on variables in the previous layer $l-1$ for readability.}
\begin{equation}\label{eq:mf1}
p(\mathbf y,\mathbf z_1,\mathbf z_2\ldots \mathbf z_L | \mathbf x)
=p(\mathbf z_1|\mathbf x)\cdot\prod_{l=2}^L p(\mathbf z_l| \mathbf z_{l-1} )\cdot p(\mathbf y|\mathbf z_L) .
\end{equation}
The probability of interest $p(\mathbf y|\mathbf x)$ is obtained by marginalization over all $\mathbf z_l$, $l=1\ldots L$. Let us first perform marginalization over values of $\mathbf z_1$ only, using thereby the approximation \eqref{eq:meanfield}. In doing so we obtain
\begin{equation}\label{eq:mf2}
p(\mathbf y,\mathbf z_2\ldots \mathbf z_L | \mathbf x)
\approx p(\mathbf z_2|\bar{\mathbf z}_1(\mathbf x))\cdot\prod_{l=3}^L p(\mathbf z_l|\mathbf z_{l-1})\cdot p(\mathbf y|\mathbf z_L) .
\end{equation}
The value of $\bar{\mathbf z}_1(\mathbf x)$ can be seen as a ``new observation'' for a network that has one layer less than the original one. By proceeding further sequentially we finally obtain
\begin{equation}\label{eq:mf3}
p(\mathbf y | \mathbf x) \approx p(\mathbf y|\bar{\mathbf z}_L(\bar{\mathbf z}_{L-1}( \ldots \bar{\mathbf z}_1(\mathbf x)))) .
\end{equation}
Hence, multi-layer FFNs perform a sequence of approximations in the corresponding BNs. We will call it {\em sequential approximation}.

At this point we would like to discuss a crucial difference between the original model \eqref{eq:mf1} and its sequential approximation \eqref{eq:mf3}. In fact, the latter implements a {\em conditionally independent} posterior probability distribution $p(\mathbf y|\mathbf x)$, since all $\bar{\mathbf z}_l$ are computed deterministically. It means, that \eqref{eq:mf3} is not able to capture the dependencies between output variables for a particular input in principle. The original model \eqref{eq:mf1} can indeed, because the desired posterior is obtained by marginalization over all hidden $\mathbf z_l$, i.e.~output variables are not conditionally independent. Hence, especially for {\em structured output} we could expect essentially better performance of the original model as compared to its approximation.


\paragraph{What is the benefit of having this connection?}\label{sec:benefits} 
In essence, the presented correspondence between FFNs and Bayesian networks allows to use for FFNs the rich and diverse set of methods that has been developed for graphical models, both directed and undirected ones, since the former can be trivially transformed into the latter. Below, we give a brief overview of possible research directions in this context.

{\em Inference}.
After training, FFNs are typically applied through forward-propagation, which we have shown to be equivalent to computing a sequential approximation \eqref{eq:mf3}. However, since we can now represent FFNs as graphical models \eqref{eq:gm}, we can try other inference strategies, such as minimizing the associated energy of \eqref{eq:gm} in order to compute the maximum a-posteriori (MAP) estimate. This task is known to be hard for BNs \cite{Saul:1996}. However, there are many elaborated techniques for energy minimization that have recently been developed for undirected graphical models \cite{Kappes2015}. Thus far, we did not pursue this direction, mainly because the corresponding energy minimization problem is of higher order, hence a specialized solver should be developed/chosen for this.

In general, we can apply Bayesian decision theory with a loss function of our choice to obtain a corresponding estimator. The associated optimization problem is typically intractable. However, note that BNs admit drawing perfect independent samples efficiently from the posterior $p(\mathbf y|\mathbf x)$. Hence, we can rather easily approximate Bayesian estimators through the use of samples. To that end, we use samples to approximate the maximum marginal decision in our experiments for binary image segmentation (see below).

{\em Learning}.
There are many approaches to learning graphical models. For example, one could use standard conditional likelihood maximization of the original model \eqref{eq:prob}.
However, we did some preliminary experiments and found this not to work well. Hence, we instead present an algorithm in the following section that maximizes a lower bound of the conditional likelihood, which we found to be quite stable and efficient. 

Another possibility would be to pose the learning task in a discriminative manner by using the framework of structural support vector machines (SSVM, see e.g.~\cite{nowozin2011structured} for an overview) with latent variables. A very important property of the SSVM framework is the possibility to learn with application-specific losses. Unfortunately, again, in order to perform this kind of learning, a specialized solver for the corresponding energy minimization problems should be developed first.

{\em CNN+CRF combinations}.
Using the representation of an FFN as a graphical model makes it essentially easier to combine a CRF with a CNN, that basically means just to extend the energy function \eqref{eq:gm} by adding the potential functions associated with the CRF.

{\em Practical benefits}.
Leveraging the connection between FFNs and the corresponding BNs gives the possibility to combine or alternate between different learning algorithms. For instance, common error back-propagation is faster as compared to BN learning. Moreover, it scales more easily to large amounts of data. Hence, a reasonable procedure could be to start learning by error back-propagation, then convert the learned FFN into the corresponding BN and perform fine-tuning by training the BN. On the other hand, stochastic methods are known to be beneficial to escape local optima. Hence, another interesting direction would be to alternate between error back-propagation and statistical BN learning, using the former to speed up convergence and the latter to escape local optima.


\section{Learning}\label{sec:learn}
We pose the learning task as maximization of the conditional data likelihood\footnote{Here we give only a brief overview of the algorithm. The complete derivation can be found in appendix \ref{appendix_b}.}
\begin{equation}\label{eq:ml}
\arg\max_{\theta}\sum_{t=1}^{|T|} \ln p(\mathbf y^t|\mathbf x^t;\theta) ,
\end{equation}
where $T=((\mathbf x^t,\mathbf y^t), t=1\ldots |T|)$ are the training data, $\theta$ summarizes all unknown parameters, and $p(\mathbf y|\mathbf x;\theta)$ is obtained by marginalization over all hidden $\mathbf z_l$, $l=1\ldots L$, organized in layers (see \eqref{eq:mf1}). In further we omit the summation over $t$ to simplify notations, i.e.~we concentrate on optimization for just one learning example $(\mathbf x,\mathbf y)$, where $\mathbf y$ is the ground truth for $\mathbf x$.

Let us assume for now that we want to learn only the parameters of the output layer. For this, we approximate \eqref{eq:ml} by its lower bound using Jensen's inequality and obtain
\begin{equation}\label{eq:jensen}
\arg\max_{\theta}\sum_{\mathbf z_L} p(\mathbf z_L|\mathbf x)\cdot \ln p(\mathbf y|\mathbf z_L;\theta),
\end{equation}
where $\mathbf z_L$ represents the last hidden layer, and $p(\mathbf z_L|\mathbf x)$ is obtained by marginalization over all hidden layers but $L$. Furthermore, let us specify $p(\mathbf y|\mathbf z_L;\theta)=\exp(f(\mathbf y,\mathbf z_L,\theta))/\sum_{\mathbf y'}\exp(f(\mathbf y',\mathbf z_L,\theta))$, where $f(\cdot)$ is some function. Using this, the gradient of \eqref{eq:jensen} wrt.~$\theta$ reads
\begin{equation}\label{eq:bla1}
\frac{\partial }{\partial \theta}=\sum_{\mathbf z_L} p(\mathbf z_L|\mathbf x)\cdot 
\left[
\frac{\partial f(\mathbf y,\mathbf z_L,\theta)}{\partial \theta}-\mathbb E_{p(\mathbf y'|\mathbf z_L;\theta)} 
\left[ \frac{\partial f(\mathbf y',\mathbf z_L,\theta)}{\partial \theta}   \right]\right] .
\end{equation}
The summations over $\mathbf z_L$ and over $\mathbf y'$ in \eqref{eq:bla1} are infeasible. Therefore, we employ {\em stochastic gradient ascent} as follows: (i) draw a sample $\hat{\mathbf z}_L$ according to the current probability distribution $p(\mathbf z_L|\mathbf x)$, (ii) draw a sample $\hat{\mathbf y}$ according to the current probability distribution $p(\mathbf y|\hat{\mathbf z}_L;\theta)$, and (iii) compute and apply the stochastic gradient
\begin{equation}\label{eq:stgrad}
\frac{\partial f(\mathbf y,\hat{\mathbf z}_L,\theta)}{\partial \theta}-
\frac{\partial f(\hat{\mathbf y},\hat{\mathbf z}_L,\theta)}{\partial \theta} .
\end{equation}
Note, that stochastic gradient ascent is not an approximation in the common sense, because it is exact in the limit of infinity. We would also like to point in that we are able to draw {\em perfect independent} samples, since we have a {\em directed} graphical model behind our probability distribution. In particular, we have to draw sequentially a $\hat{\mathbf z}_1$ from $p(\mathbf z_1|\mathbf x)$, then $\hat{\mathbf z}_2$ from $p(\mathbf z_2|\hat{\mathbf z}_1)$ and so on up to the $\hat{\mathbf z}_L$ needed in \eqref{eq:stgrad}. Note also, that each sampling step can be performed very efficiently, since all probability distributions $p(\mathbf z_l|\mathbf z_{l-1})$ are conditionally independent. This is in contrast to sampling in general undirected graphical models using e.g.~Gibbs sampling, where obtaining independent samples is quite time consuming.

In order to learn the parameters of an $l$-th layer, we proceed as follows. We ``split'' the network in two parts -- one from $\mathbf x$ to $\mathbf z_{l-1}$ and another from $\mathbf z_l$ to the output $\mathbf y$. Again, using Jensen's inequality, the corresponding lower bound of the likelihood \eqref{eq:ml} reads
\begin{equation}\label{eq:jensen1}
\arg\max_{\theta}\sum_{\mathbf z_{l-1}} p(\mathbf z_{l-1}|\mathbf x)\cdot \ln p(\mathbf y|\mathbf z_{l-1};\theta) .
\end{equation}
The summation over $\mathbf z_{l-1}$ is again carried out using stochastic approximation. For $\ln p(\mathbf y|\mathbf z_{l-1};\theta)$ we use the approximation \eqref{eq:mf3}, i.e.~we approximate the part of BN from $\mathbf z_l$ to $\mathbf y$ by the corresponding FFN. 

To summarize, the resulting algorithm consists of two steps: (i) draw a sample $\hat{\mathbf z}_1, \hat{\mathbf z}_2\ldots \hat{\mathbf z}_L, \hat{\mathbf y}$ using current model, and (ii) propagate an ``error'' through the network, compute and apply stochastic gradient, like \eqref{eq:stgrad} for the output layer. The second step is thereby exactly the same as in the commonly used error back-propagation (EBP) for FFNs. The difference is only in the first step -- EBP deterministically computes the output value for each neuron, whereas in our algorithm the output values are sampled according to the corresponding conditional probability distributions. In essence, the core difference between EBP and our statistical learning is the following. EBP {\em approximates} the underlying probability distribution but uses {\em exact} gradient. Our approach is to train the {\em original} model, but use {\em approximated} gradient ascent for learning.


\section{Experiments}
\paragraph{Image classification.} Here, we validate our approach on the CIFAR-10 dataset \cite{Krizhevsky09learningmultiple}\footnote{The detail about the dataset can be found under \url{https://www.cs.toronto.edu/~kriz/cifar.html}.}. We used Caffe framework \cite{jia2014caffe} in these experiments and the ``standard'' network implementation\footnote{\url{http://caffe.berkeleyvision.org/gathered/examples/cifar10.html}.} as the baseline. First, we experimented with the network variant built of sigmoid neurons. In order to further simplify experiments and avoid influences of additional design choices we used simple stochastic gradient descent solver without momentum and fixed gradient step size. At the same time we slightly increased the step size and learned models with essentially higher number of iterations. Both EBP for the baseline FFN and our algorithm for the corresponding BN were learned with exactly the same parameters. We also performed several runs for both methods and averaged the measurements.

\begin{figure}[h]
\begin{center}
\begin{minipage}{0.33\textwidth}
\includegraphics[width=\textwidth]{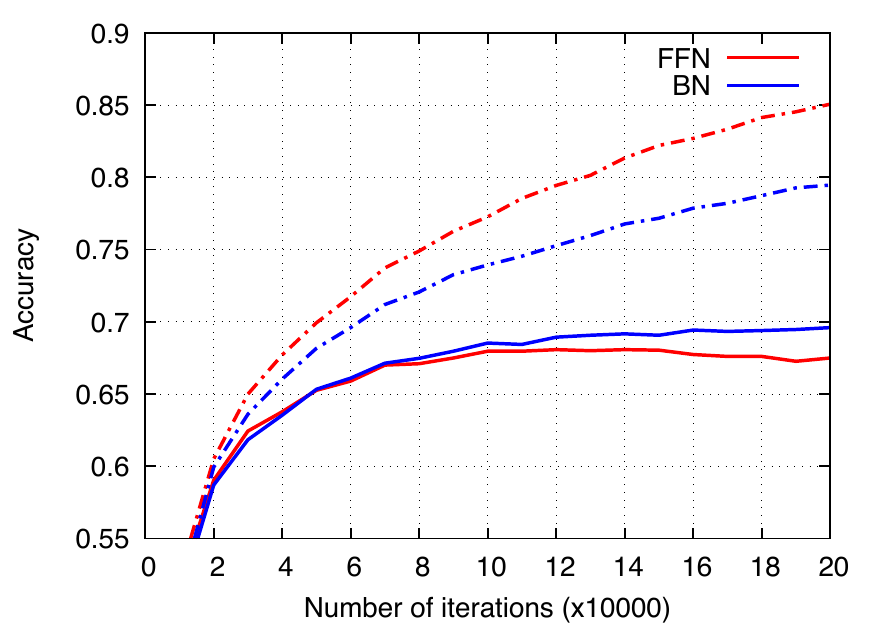}
\end{minipage}\hfill
\begin{minipage}{0.66\textwidth}
\begin{center}
\begin{tabular}{|l|c|c|c|}
\hline 
 & Training & ~~~~Test~~~~ & Difference \\
\hline 
FFN, sigmoids & 85.1 & 67.5 & 17.6 \\
BN, sigmoids & 79.5 & {\bf 69.6} & {\bf 9.9} \\
\hline
FFN, RELUs (orig.) & 95.0 & {\bf 74.3} & 20.7 \\
FFN, RELUs (appr.) & 79.3 & 71.5 & 7.8 \\
BN, RELUs & 79.1 & 72.6 & {\bf 6.5} \\
\hline
\end{tabular}
\end{center}
\end{minipage}
\end{center}
\caption{\label{fig:cifar10}Results for CIFAR-10 benchmark. Left -- training and test accuracies for the baseline and the proposed approach (with sigmoid neurons) as the dependencies on time during learning, dashed lines -- training, solid lines -- test. Right -- results summary (classification accuracies in \%), ``orig.'' -- the original network, ``appr.'' -- RELUs are approximated by the sum of sigmoids as in \cite{nair2010rectified}.}
\end{figure}

Fig.~\ref{fig:cifar10} (left) shows the dependencies of the training and test accuracies on time. At the beginning of the learning process the test accuracy of the FFN slightly outperforms the one of the BN. After relatively short time however it saturates and even starts to decrease, that is a sign for over-fitting. In contrast, the BN's test accuracy constantly increases and finally outperforms FFN by about 2\%. What is however even more important in our opinion is that the BN shows essentially smaller {\em difference} between the training and the test accuracies. Hence, it is much more robust to over-fitting as compared to its deterministic counterpart. The results are summarized in Fig.~\ref{fig:cifar10} (right).

We also experimented with the RELU-variant of the network. In doing so, we used all default parameters (step size, momentum, weight decay etc.), increasing only the number of learning iterations. Remember that first we need to approximate RELUs by a sum of sigmoid neurons. Actually, it makes the test accuracy essentially worse (see Fig.~\ref{fig:cifar10} (right), ``orig.'' vs.~``appr.''). On the other side, the original RELU network turns out to be extremely over-fitting. As in the previous esperiment with sigmoids, BN improves both the performance and the generalization of its deterministic counterpart. Among all experiments, the BN with stochastic RELUs shows the best generalization.

\paragraph{Binary image segmentation.} Next, we evaluate the proposed method for binary image segmentation on Weizmann Horse dataset \cite{borenstein2004combining}. It consists of 328 images, showing horses in a side view on different backgrounds. In order to accelerate computations, we downscaled all images to the size of $80{\times}80$ pixels.

\begin{figure}
\begin{center}
\begin{minipage}{0.333\textwidth}
\begin{center}
\includegraphics[width=\textwidth]{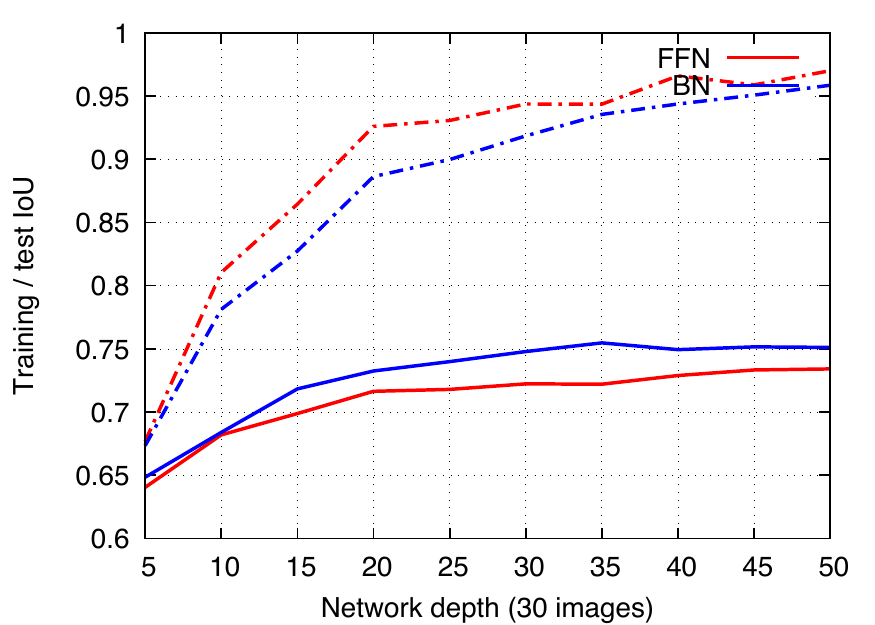}
\end{center}
\end{minipage}\hfill
\begin{minipage}{0.333\textwidth}
\begin{center}
\includegraphics[width=\textwidth]{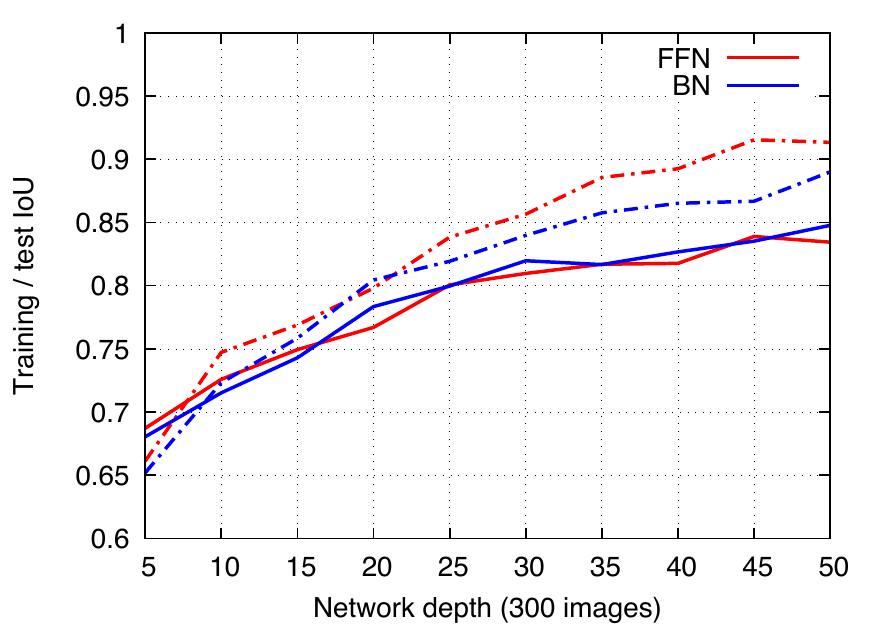}
\end{center}
\end{minipage}\hfill
\begin{minipage}{0.333\textwidth}
\begin{center}
\includegraphics[width=\textwidth]{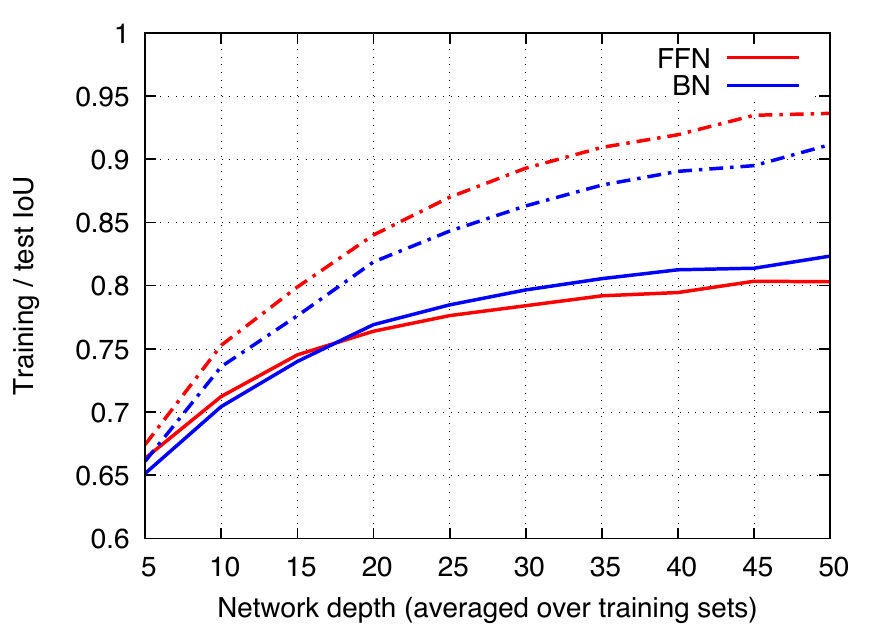}
\end{center}
\end{minipage}

\begin{minipage}{0.333\textwidth}
\begin{center}
\vspace{1ex}
\scalebox{0.82}{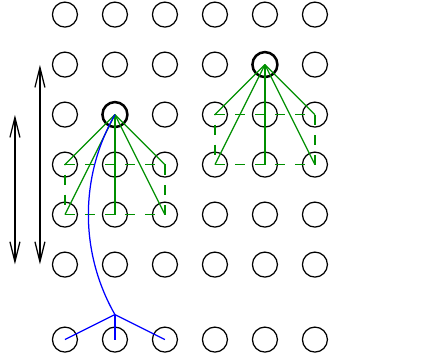}
\end{center}
\end{minipage}\hfill
\begin{minipage}{0.333\textwidth}
\begin{center}
\includegraphics[width=\textwidth]{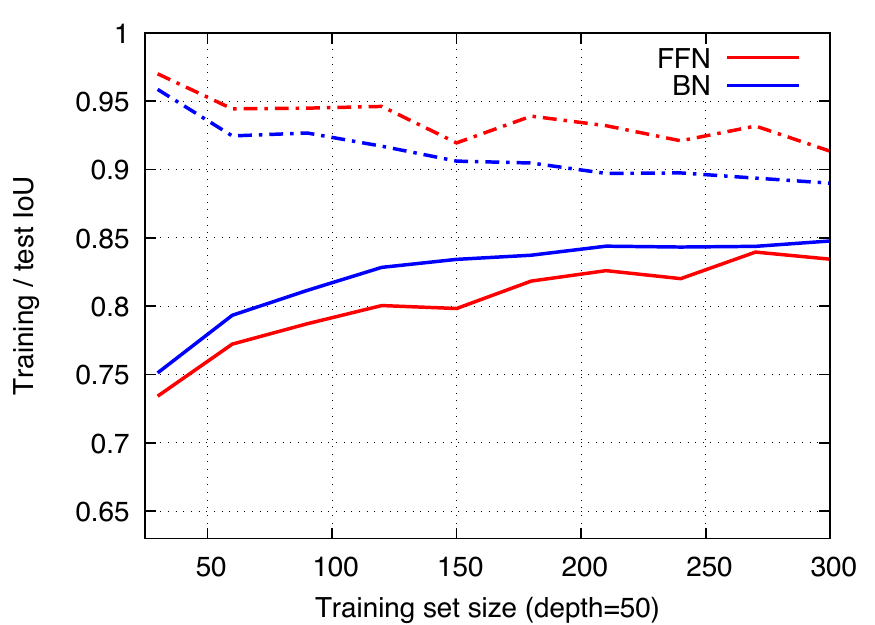}
\end{center}
\end{minipage}\hfill
\begin{minipage}{0.333\textwidth}
\begin{center}
\includegraphics[width=\textwidth]{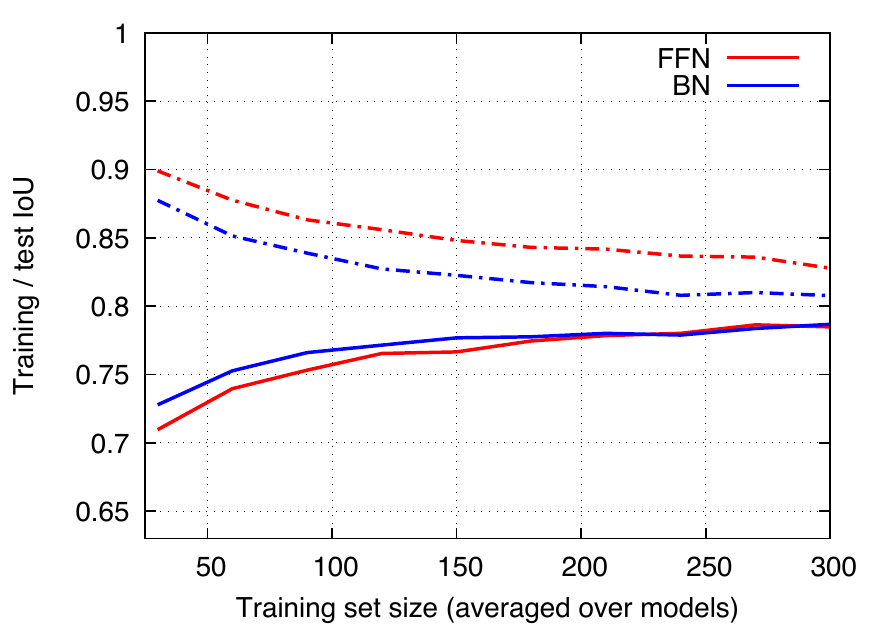}
\end{center}
\end{minipage}
\end{center}
\caption{\label{fig:traintest}Results for Weizmann Horse dataset. Dependencies of the IoU on the model capacity (top row) and on the size of the training set (bottom row, middle and right). Top, left -- weakest training set, middle -- strongest training set/model, right -- averages over training sets/models. Dashed lines -- training, solid lines -- test. Bottom, left -- the used network architecture (only one image row is shown). The input sets for two neurons depicted in bold are shown. The connection to intermediate variables are shown in green, connections to the image -- in blue. For this example, $\delta_l=2$ and $\delta_r=1$.}
\end{figure}

For this experiment we designed our own convolutional (CNN) architecture shown in Fig.~\ref{fig:traintest} (left, bottom). Formally, the input variables $v\in X$ correspond to image pixels, whereas $x_v\in \mathbb R^3$ is the RGB-color. There is one output variable $v\in Y$ for each pixel that assigns a label $y_v\in\{0,1\}$ to it. Neurons are organized in layers with one neuron for each pixel. We will refer all neurons by their ``coordinates'' $(l,i,j)$, where $l$ is the layer number and $(i,j)$ are image coordinates. The set of inputs $I_v=I_{(l,i,j)}$ for each neuron consists of neurons from $\delta_l$ previous layers that are within a certain radius $\delta_r$ around $(i,j)$. Neurons from the first $L'$ layers have also access to the image pixels, again within the radius $\delta_r$. All neurons are sigmoids.

We study the model performance as well as the generalization capability with respect to varying network depth and the number of training examples\footnote{Further experiments can be found in appendix \ref{appendix_c}.}. One experiment consists in following. First, we draw randomly a pre-defined number of training examples (out of 328 in total), the rest is used for test. Then we learn the deterministic variant of the CNN by EBP and the corresponding BN by the proposed algorithm. After the models are learned we perform inference using maximum marginal decision strategy and compute intersection over union (IoU) for both models and both training and test sets. Each experiment is repeated 10 times (i.e.~with different randomly sampled training/test sets of the same size) and the measures are averaged. 

Fig.~\ref{fig:traintest} shows the dependencies of IoU on the network depth (upper row) and on the training set size (bottom row, middle and right). First of all, it is clearly seen that statistical model has essentially better generalization capabilities, since the difference between training and test accuracies is always smaller. Concerning performance (i.e.~IoU on the test set), statistical model is also superior in most cases. This performance gain decreases for larger training sets (see Fig.~\ref{fig:traintest}, top, middle, and bottom, right). This is indeed expected, since it is well known that discriminative approaches are less robust with respect to over-fitting in general, as compared to the statistical ones. At the same time, they are less vulnerable to misspecification, i.e.~they perform better for large training data in general.

\section{Conclusion}
In this work we established a connection between feed-forward neural networks and Bayesian networks. We showed that the application of an FFN can be seen as an approximation of the corresponding BN. Based on this connection, we developed a learning algorithm for BNs. Evaluation shows that BNs outperform the corresponding FFNs in terms of both performance and generalization capabilities. We also characterized various benefits of having this connection between FFNs and GMs (see the discussion in Sec.~\ref{sec:benefits}), which motivate us for further work.


{\small
\bibliographystyle{plain}
\bibliography{arxiv}
}


\begin{appendices}

\section{Approximation with discrete variables.}\label{appendix_a}
First of all we would like to discuss in more detail the approximation 
\begin{equation}\label{eq_app:meanfield}
p(\mathbf y|\mathbf x)=\sum_{\mathbf z} p(\mathbf y|\mathbf z)\cdot p(\mathbf z|\mathbf x)\approx p(\mathbf y|\bar{\mathbf z}(\mathbf x)) ,
\text{\ where\ \ \ } \bar{\mathbf z}(\mathbf x)=\sum_{\mathbf z} \mathbf z\cdot p(\mathbf z|\mathbf x) .
\end{equation}
(see \eqref{eq:meanfield} in the main part), especially for the case of discrete variables $\mathbf z$. 

Consider e.g.~the case that the variable $z$ of some conditional probability distribution $p(z|x)$ is discrete, e.g.~$z \in \{a,b\}$, where $a$ and $b$ are not numbers, but just elements of a general discrete set (like labels). The expectation $\bar{z}(x)=\mathbb E_{p(z|x)}[z]$ is not defined in this case, because addition and multiplication are not defined. In praxis however, the values of $z$ are supposed to be (although discrete) elements of a vector space in fact, i.e.~they are ``encoded'' e.g.~by $\{0,1\}$, $\{-1,+1\}$, or similar. Moreover, often the values of a random discrete variable $\mathbf z$ are further used in another probability distribution $p(y|\mathbf z)$, where $p(y|\mathbf z)$ is specified by scalar product, i.e.~$p(y|\mathbf z)=p(y|\langle\mathbf z,w\rangle)$. Hence again, the values of $\mathbf z$ should be vectors in order to be able to define scalar product.

In order to be able to work with discrete variables, let us explicitly introduce the ``encoding function'' $\xi(\cdot)$, the argument of which is a value of a random variable, while output is an element of a vector space, i.e.~addition and multiplication with a scalar are well-defined for values of $\xi$. Using the encoding function makes possible to consider the expectation of values of $\xi$ instead of values of $z$, i.e.~$\mathbb E_{p(y|x)}[\xi]$, which is indeed well-defined. For instance, in the example above, $a$ and $b$ are to be considered as {\em events}, whereas $\xi:\{a,b\}\rightarrow\mathbb R$ is a {\em random variable}. For instance, $\xi(a) = 0$ and $\xi(b) = 1$ is the encoding used in sigmoid neurons, $\xi(a) = -1$ and $\xi(b) = 1$ corresponds to a $\tanh$-neuron, etc. Using these notations \eqref{eq_app:meanfield} can be rewritten as
\begin{equation}\label{eq_app:meanfieldxi}
p(\mathbf y|\mathbf x)=\sum_{\mathbf z} p(\mathbf y|\xi(\mathbf z))\cdot p(\mathbf z|\mathbf x)\approx p(\mathbf y|\bar{\xi}(\mathbf x)) ,
\text{\ where\ \ \ } \bar{\xi}(\mathbf x)=\sum_{\mathbf z} \xi(\mathbf z)\cdot p(\mathbf z|\mathbf x) .
\end{equation}
Correspondingly, the {\em sequential approximation} 
\begin{equation}\label{eq_app:mf3}
p(\mathbf y | \mathbf x) \approx p(\mathbf y|\bar{\mathbf z}_L(\bar{\mathbf z}_{L-1}( \ldots \bar{\mathbf z}_1(\mathbf x)))) 
\end{equation}
(see \eqref{eq:mf3} in the main part) can be rewritten as 
\begin{equation}\label{eq_app:mf3xi}
p(\mathbf y | \mathbf x) \approx p(\mathbf y|\bar{\xi}_L(\bar{\xi}_{L-1}( \ldots \bar{\xi}_1(\mathbf x)))) .
\end{equation}
In further however, we will still use ``simplified'' notations $\bar{\mathbf z}$ instead of ''mathematically correct'' $\bar{\xi}$ just to omit notational clutter.


\section{The learning algorithm.}\label{appendix_b}
Here, we give a more detailed derivation of our learning algorithm. Let us first recall the problem statement. Our model is (see \eqref{eq:mf1} in the main part)
\begin{equation}\label{eq_app:mf1}
p(\mathbf y,\mathbf z_1,\mathbf z_2\ldots \mathbf z_L | \mathbf x)=
p(\mathbf z_1|\mathbf x)\cdot\prod_{l=2}^L p(\mathbf z_l| \mathbf z_{l-1} )\cdot p(\mathbf y|\mathbf z_L) .
\end{equation}
The learning task is posed as maximization of the conditional data likelihood
\begin{equation}\label{eq_app:ml}
\arg\max_{\theta}\sum_{t=1}^{|T|} \ln p(\mathbf y^t|\mathbf x^t;\theta) ,
\end{equation}
where $T=((\mathbf x^t,\mathbf y^t), t=1\ldots |T|)$ are the training data, $\theta$ summarizes all unknown parameters, and $p(\mathbf y|\mathbf x;\theta)$ is obtained by marginalization of \eqref{eq_app:mf1} over all hidden $\mathbf z_l$, $l=1\ldots L$. In further we omit the summation over $t$ to simplify notations, i.e.~we concentrate on optimization for just one learning example $(\mathbf x,\mathbf y)$, where $\mathbf y$ is the ground truth for $\mathbf x$.

\paragraph{Output layer.} Let us assume for now that we want to learn only the parameters of the output layer. For this, we approximate \eqref{eq_app:ml} by its lower bound using Jensen's inequality and obtain
\begin{equation}\label{eq_app:jensen}
\arg\max_{\theta}\sum_{\mathbf z_L} p(\mathbf z_L|\mathbf x)\cdot \ln p(\mathbf y|\mathbf z_L;\theta),
\end{equation}
where $\mathbf z_L$ represents the last hidden layer, and $p(\mathbf z_L|\mathbf x)$ is obtained by marginalization over all hidden layers but $L$. Furthermore, let us specify $p(\mathbf y|\mathbf z_L;\theta)=\exp(f(\mathbf y,\mathbf z_L,\theta))/\sum_{\mathbf y'}\exp(f(\mathbf y',\mathbf z_L,\theta))$, where $f(\cdot)$ is some function. Using this, the gradient of \eqref{eq_app:jensen} reads
\begin{equation}\label{eq_app:bla1}
\frac{\partial }{\partial \theta}=\sum_{\mathbf z_L} p(\mathbf z_L|\mathbf x)\cdot 
\left[
\frac{\partial f(\mathbf y,\mathbf z_L,\theta)}{\partial \theta}-\mathbb E_{p(\mathbf y'|\mathbf z_L;\theta)} 
\left[ \frac{\partial f(\mathbf y',\mathbf z_L,\theta)}{\partial \theta}   \right]\right] .
\end{equation}
The summations over $\mathbf z_L$ and over $\mathbf y'$ in \eqref{eq_app:bla1} are infeasible. Therefore, we employ {\em stochastic gradient ascent}, i.e.~we replace the {\em expectation} over $\mathbf z_L$ by a {\em realization}. In particular, first, we draw a sample $\hat{\mathbf z}_L$ according to the current probability distribution $p(\mathbf z_L|\mathbf x)$, and second, apply stochastic gradient 
\begin{equation}\label{eq_app:bla1a}
\frac{\partial f(\mathbf y,\hat{\mathbf z}_L,\theta)}{\partial \theta}-\mathbb E_{p(\mathbf y'|\hat{\mathbf z}_L;\theta)} 
\left[ \frac{\partial f(\mathbf y',\hat{\mathbf z}_L,\theta)}{\partial \theta}   \right] .
\end{equation}
Moreover, the second addend in \eqref{eq_app:bla1a} is again an expectation. Hence, we can employ stochastic approximation here as well. In particular, given an $\hat{\mathbf z}_L$, we draw an $\hat{\mathbf y}$ from $p(\mathbf y|\hat{\mathbf z}_L;\theta)$, and our final stochastic gradient is
\begin{equation}\label{eq_app:stgrad}
\frac{\partial f(\mathbf y,\hat{\mathbf z}_L,\theta)}{\partial \theta}-
\frac{\partial f(\hat{\mathbf y},\hat{\mathbf z}_L,\theta)}{\partial \theta} .
\end{equation}

To be more concrete, let us consider an example, where the output layer consists of sigmoid neurons, i.e.
\begin{equation}\label{eq_app:sigms}
f(y_v,\mathbf z,w_v)=y_v\cdot\langle\mathbf z,w_v\rangle ,
\end{equation}
where the lower index $v$ indicates a particular output neuron, and $w_v$ represents its weights. Correspondingly, the gradient \eqref{eq_app:stgrad} is
\begin{equation}
y_v\cdot \hat{\mathbf z}_L - \hat{y}_v\cdot \hat{\mathbf z}_L=\hat{\mathbf z}_L\cdot(y_v-\hat{y}_v) .
\end{equation}


\paragraph{Intermediate layers.} In order to learn the parameters of an $l$-th layer, we proceed as follows. We ``split'' the network in two parts: one from $\mathbf x$ to $\mathbf z_{l-1}$ and another from $\mathbf z_{l-1}$ to the output $\mathbf y$. Again, using Jensen's inequality, the corresponding lower bound of the likelihood \eqref{eq_app:ml} reads
\begin{equation}\label{eq_app:jensen1}
\arg\max_{\theta}\sum_{\mathbf z_{l-1}} p(\mathbf z_{l-1}|\mathbf x)\cdot 
\ln p(\mathbf y|\mathbf z_{l-1};\theta) .
\end{equation}
The conditional probability distribution $p(\mathbf y|\mathbf z_{l-1};\theta)$ is obtained by marginalization over all layers $\mathbf z_{l'}$, $l'=l\ldots L$. The key idea to cope with this infeasible marginalization is to approximate $p(\mathbf y|\mathbf z_{l-1};\theta)$ by the corresponding feed-forward network using our approximation \eqref{eq_app:mf3}, i.e.
\begin{equation*}
p(\mathbf y|\mathbf z_{l-1};\theta)\approx
p(\mathbf y|\bar{\mathbf z}_L(\bar{\mathbf z}_{L-1}( \ldots \bar{\mathbf z}_l(\mathbf z_{l-1},\theta)))) ,
\end{equation*}
which will be denoted by $\tilde p (\mathbf y|\bar{\mathbf z}_l(\mathbf z_{l-1},\theta))$. After this approximation the lower bound \eqref{eq_app:jensen1} reads
\begin{equation}\label{eq_app:mlapprox}
\arg\max_{\theta}\sum_{\mathbf z_{l-1}} p(\mathbf z_{l-1}|\mathbf x) \cdot
\ln \tilde p(\mathbf y|\bar{\mathbf z}_l(\mathbf z_{l-1},\theta)) ,
\text{\ where\ \ } \bar{\mathbf z}_l(\mathbf z_{l-1},\theta)= \sum_{\mathbf z_l} \mathbf z_l\cdot p(\mathbf z_l|\mathbf z_{l-1};\theta).
\end{equation}
Its gradient is w.r.t.~$\theta$ is obtained via chain rule as
\begin{equation}
\sum_{\mathbf z_{l-1}} p(\mathbf z_{l-1}|\mathbf x) \cdot
\frac{\partial \ln \tilde p(\mathbf y|\bar{\mathbf z}_l(\mathbf z_{l-1},\theta))}{\partial \bar{\mathbf z}_l(\mathbf z_{l-1},\theta)}\cdot
\frac{\partial \bar{\mathbf z}_l(\mathbf z_{l-1},\theta)}{\partial\theta} .
\end{equation}
The summation over $\mathbf z_{l-1}$ is again carried out using stochastic approximation: (i) draw $\hat{\mathbf z}_{l-1}$ from $p(\mathbf z_{l-1}|\mathbf x)$, and (ii) apply the stochastic gradient
\begin{equation}\label{eq_app:stgradl}
\frac{\partial \ln \tilde p(\mathbf y|\bar{\mathbf z}_l(\hat{\mathbf z}_{l-1},\theta))}{\partial \bar{\mathbf z}_l(\hat{\mathbf z}_{l-1},\theta)}\cdot
\frac{\partial \bar{\mathbf z}_l(\hat{\mathbf z}_{l-1},\theta)}{\partial\theta} .
\end{equation}
The first factor in \eqref{eq_app:stgradl} is obtained using standard error back-propagation. It computes an {\em error} $\mathbf \delta_l$, which is a vector of values $\delta_{lv}$ for each neuron $v$ in the $l$-th layer. The second factor in \eqref{eq_app:stgradl} is
\begin{equation}\label{eq_app:stgradl1}
\frac{\partial \bar{\mathbf z}_l(\hat{\mathbf z}_{l-1},\theta)}{\partial\theta}=
\frac{\partial}{\partial\theta}\left[
\sum_{\mathbf z_l}\mathbf z_l\cdot p(\mathbf z_l|\hat{\mathbf z}_{l-1};\theta)
\right]=
\sum_{\mathbf z_l}\mathbf z_l\cdot
\frac{\partial p(\mathbf z_l|\hat{\mathbf z}_{l-1};\theta)}{\partial f(\mathbf z_l,\hat{\mathbf z}_{l-1},\theta)}\cdot
\frac{\partial f(\mathbf z_l,\hat{\mathbf z}_{l-1},\theta)}{\partial \theta}.
\end{equation}
Combining \eqref{eq_app:stgradl} and \eqref{eq_app:stgradl1} for a particular neuron $v$ we obtain the stochastic gradient w.r.t.~its parameters $\theta_v$ as
\begin{equation}\label{eq_app:stgradl2}
\delta_{lv}\cdot\sum_{z_v} z_v\cdot
\frac{\partial p(z_v|\hat{\mathbf z}_{l-1};\theta_v)}{\partial f(z_v,\hat{\mathbf z}_{l-1},\theta_v)}\cdot
\frac{\partial f(z_v,\hat{\mathbf z}_{l-1},\theta_v)}{\partial \theta_v}.
\end{equation}

Let us again concretize \eqref{eq_app:stgradl2} for sigmoid neurons \eqref{eq_app:sigms}. Remember that in this case $z_v\in\{0,1\}$, i.e.~the sum over $z_v$ in \eqref{eq_app:stgradl2} is reduced to just one value for $z_v=1$. Hence, the stochastic gradient for sigmoid neurons is
\begin{eqnarray}
& & \delta_{lv}\cdot\frac{\partial p(z_v=1|\hat{\mathbf z}_{l-1};w_v)}{\partial f(z_v=1,\hat{\mathbf z}_{l-1},w_v)}\cdot
\frac{\partial f(z_v=1,\hat{\mathbf z}_{l-1},w_v)}{\partial w_v}= \nonumber\\
& & = \delta_{lv}\cdot\frac{\exp(a)}{(1+\exp(a))^2}\cdot \hat{\mathbf z}_{l-1} ,
\text{\ with \ \ } a=\langle\hat{\mathbf z}_{l-1},w_v\rangle .
\end{eqnarray}

There is a practical concern with the resulting algorithm. The problem is that for each layer we need to perform its own forward and backward pass to compute the first factor in \eqref{eq_app:stgradl}, that is quite time consuming. Therefore in practice, we perform only one forward and one backward pass per gradient update to save computations. During the forward pass we sample values $\hat{z}_{lv}$ and $\hat{y}_v$, which are necessary for stochastic approximations \eqref{eq_app:bla1a}, \eqref{eq_app:stgrad} and \eqref{eq_app:stgradl}. During the backward pass we use these sampled values for error propagation and update parameters by \eqref{eq_app:stgrad} and \eqref{eq_app:stgradl2}.

All in all, the resulting algorithm turns out to be very similar to the commonly used error back-propagation. The difference is only in the forward pass -- error back-propagation {\em deterministically} computes the output value for each neuron, whereas in our algorithm the output values are {\em sampled} according to the corresponding conditional probability distribution.


\section{Additional experiments}\label{appendix_c}

Here we present some additional experiments for binary image segmentation on Weizmann Horse dataset.

\paragraph{Robustness against image noise.}
\begin{figure}
\begin{center}
\includegraphics[width=0.197\linewidth]{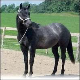}\hfill
\includegraphics[width=0.197\linewidth]{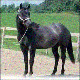}\hfill
\includegraphics[width=0.197\linewidth]{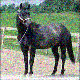}\hfill
\includegraphics[width=0.197\linewidth]{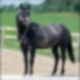}\hfill
\includegraphics[width=0.197\linewidth]{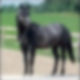}

\includegraphics[width=0.197\linewidth]{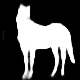}\hfill
\includegraphics[width=0.197\linewidth]{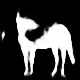}\hfill
\includegraphics[width=0.197\linewidth]{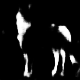}\hfill
\includegraphics[width=0.197\linewidth]{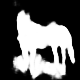}\hfill
\includegraphics[width=0.197\linewidth]{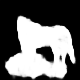}

\includegraphics[width=0.197\linewidth]{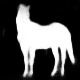}\hfill
\includegraphics[width=0.197\linewidth]{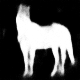}\hfill
\includegraphics[width=0.197\linewidth]{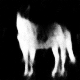}\hfill
\includegraphics[width=0.197\linewidth]{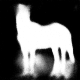}\hfill
\includegraphics[width=0.197\linewidth]{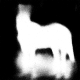}
\end{center}
\caption{\label{fig:noise}Results for images with increasing noise level. Middle row -- FFN, bottom row -- BN. Left -- original image, second and third columns -- Gaussian noise of increasing variance, fourth and fifth columns -- Gaussian blur with increasing radius.}
\end{figure}
This experiment aims to compare the behaviour of the models with respect to image noise. For this, we learned two models: a feed-forward network (FFN) and the corresponding Bayesian network (BN) with depth equal 50 on the whole dataset of 328 images (see the model description in the experimental section of the main part). Perhaps the most popular noise types are pixel-wise independent Gaussian noise and Gaussian blur. We performed series of experiments with an image from the training set, which was segmented very well by both FFN and BN. We corrupted it by (i) Gaussian noise of increasing variance and (ii) Gaussian blur with increasing radius. The results are shown in Fig.~\ref{fig:noise}. We prefer to give not the decisions but gray-value coded foreground marginals, since these are more illustrative in our opinion. It can be clearly seen that the segmentations produced by FFN degenerate faster with the increasing noise level as compared to those produced by the corresponding BN. It asserts the favourable robustness of the latter to image noise.

\paragraph{Semi-supervised segmentation.}
\begin{figure}
\begin{center}
\includegraphics[width=0.33\linewidth]{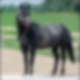}\hfill
\includegraphics[width=0.33\linewidth]{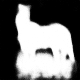}\hfill
\includegraphics[width=0.33\linewidth]{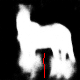}
\end{center}
\caption{\label{fig:post} Results for interactive segmentation. Middle -- without and right -- with user scribbles (marked red).}
\end{figure}
Let us assume that at the inference stage some pixels are marked by the user as being the foreground or background. Obviously, this additional information (we call it ``user scribbles'') does not play any role, when working with FFN, again, because the underlying probability distribution for output variables is {\em conditionally independent} given an image. Hence, a given decision for a particular pixel does not influence other pixels at all. In the corresponding BN in contrast, the additional user information can be taken into account as follows. Let us denote by $Y_u\subset Y$ the subset of output variables that are set by user, and $\mathbf y_u$ the vector of corresponding values. Let $Y_d\subset Y$ be the set of all other output variables, and $\mathbf y_d$ the vector of their values. Instead to consider $p(\mathbf y|\mathbf x)$ as at the usual inference, one can consider the posterior $p(\mathbf y_d|\mathbf x,\mathbf y_u)$. In order to get the decision according to the maximum marginal strategy, we need marginal probability distributions for unknown output variables $p(y_v|\mathbf x,\mathbf y_u)$ for $v\in Y_d$, which are obtained by marginalization over all hidden $\mathbf z$-s. Since we have a usual graphical model (although of higher order, see \eqref{eq:gm} in the main paper), we can estimate these probabilities. In particular, we sample configurations according to the posterior $p(\mathbf y_d, \mathbf z_1\ldots\mathbf z_L|\mathbf x,\mathbf y_u)$ using Gibbs sampling. The results are presented in Fig.~\ref{fig:post}. We used an ``erroneously'' segmented image from the previous experiment with blur. Note, how greatly the user scribble influences the marginal distributions even in pixels that are quite far away from it.

\end{appendices}


\end{document}